\documentclass{egpubl}
\usepackage{pg2018}

 \SpecialIssuePaper         
 \electronicVersion 

\ifpdf \usepackage[pdftex]{graphicx} \pdfcompresslevel=9
\else \usepackage[dvips]{graphicx} \fi

\PrintedOrElectronic

\usepackage{t1enc,dfadobe}

\usepackage{egweblnk}
\usepackage{cite}
\usepackage[none]{hyphenat}

\usepackage{bm}
\usepackage{mathptmx}
\usepackage{times}
\usepackage{color}
\usepackage{amssymb}
\usepackage{amsmath}
\usepackage{algorithm}
\usepackage{algorithmic}

\usepackage{epstopdf}
\usepackage{multirow}
\usepackage{subfig}
\usepackage{float}
\usepackage{arydshln}

\DeclareMathOperator*{\argmin}{arg\,min}
\usepackage{pdfpages}
\title[Binocular Tone Mapping with Improved Overall Contrast and Local Details]%
      {Binocular Tone Mapping with Improved Overall Contrast and Local Details}

\author[Z. Zhang \& X. Hu \& X. Liu \& T. Wong]
{\parbox{\textwidth}{\centering Zhuming Zhang$^{2,1}$, Xinghong Hu$^{2,1}$,  Xueting Liu $^{1}$ and Tien-Tsin Wong\thanks{Corresponding author: Tien-Tsin Wong (ttwong@cse.cuhk.edu.hk).}$^{1,2}$
	}
	\\
	{\parbox{\textwidth}{\centering $^1$The Chinese University of Hong Kong\\
			$^2$Shenzhen Key Laboratory of Virtual Reality and Human Interaction Technology, Shenzhen Institutes of Advanced Technology,\\ Chinese Academy of Sciences, China
		}
	}
}

\begin{document}

\maketitle
\begin{abstract}
	Tone mapping is a commonly used technique that maps the set of colors in high-dynamic-range (HDR) images to another set of colors in low-dynamic-range (LDR) images, to fit the need for print-outs, LCD monitors and projectors. Unfortunately, during the compression of dynamic range, the overall contrast and local details generally cannot be preserved simultaneously. Recently, with the increased use of stereoscopic devices, the notion of binocular tone mapping has been proposed in the existing research study. However, the existing research lacks the binocular perception study and is unable to generate the optimal binocular pair that presents the most visual content. In this paper, we propose a novel perception-based binocular tone mapping method, that can generate an optimal binocular image pair (generating left and right images simultaneously) from an HDR image that presents the most visual content by designing a binocular perception metric. Our method outperforms the existing method in terms of both visual and time performance.


\end{abstract}  

\section{Introduction}
\label{sec:intro}

\begin{figure}[!t]
	\centering
	\includegraphics[width=\linewidth]{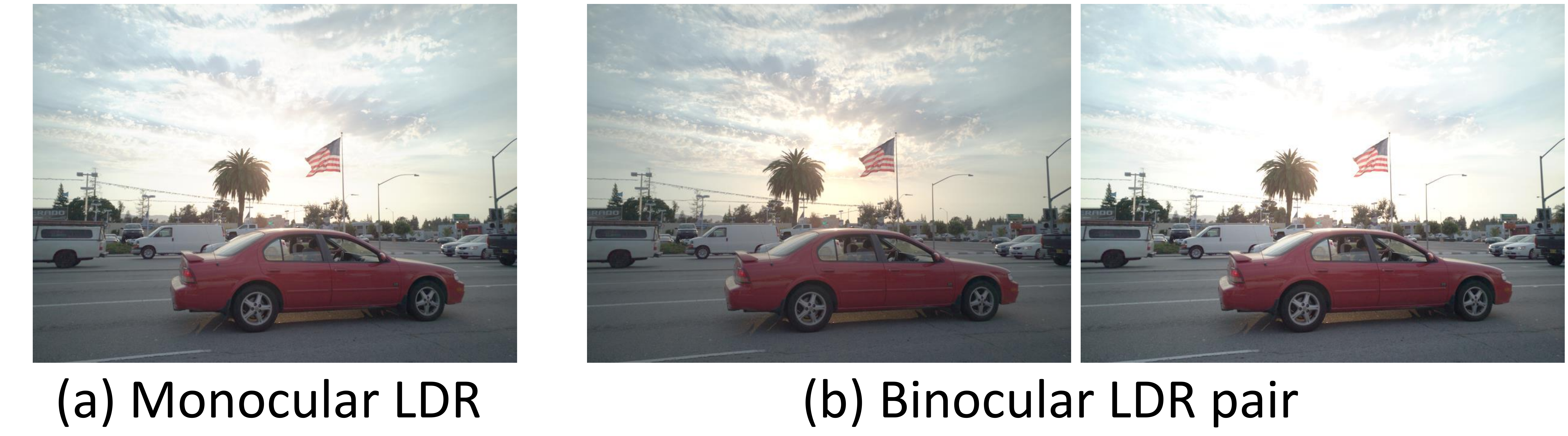}
	\caption{Monucular LDR vs. Binocular LDR pair.}
	\label{fig:intro_pair_lrmean}
\end{figure}

Tone mapping is a commonly used technique that maps the set of colors in high-dynamic-range (HDR) images to another set of colors in low-dynamic-range (LDR) images, to fit the need for print-outs, LCD monitors and projectors. Unfortunately, tone mapping leads to unavoidable information loss due to the reduced dynamic range. During the compression of dynamic range, the overall contrast and local details generally cannot be preserved simultaneously (Fig.~\ref{fig:intro_pair_lrmean}(a))~\cite{rizzi2004human}. To preserve overall contrast, details in extremely dark (minimal illumination) or extremely bright (maximal illumination) regions would be lost (the right image of Fig.~\ref{fig:intro_pair_lrmean}(b)). To preserve the details in extreme-illumination regions, the image would look ``flat'' (the left image of Fig.~\ref{fig:intro_pair_lrmean}(b)).

\begin{figure}[!htbp]
\centering
\includegraphics[width=\linewidth]{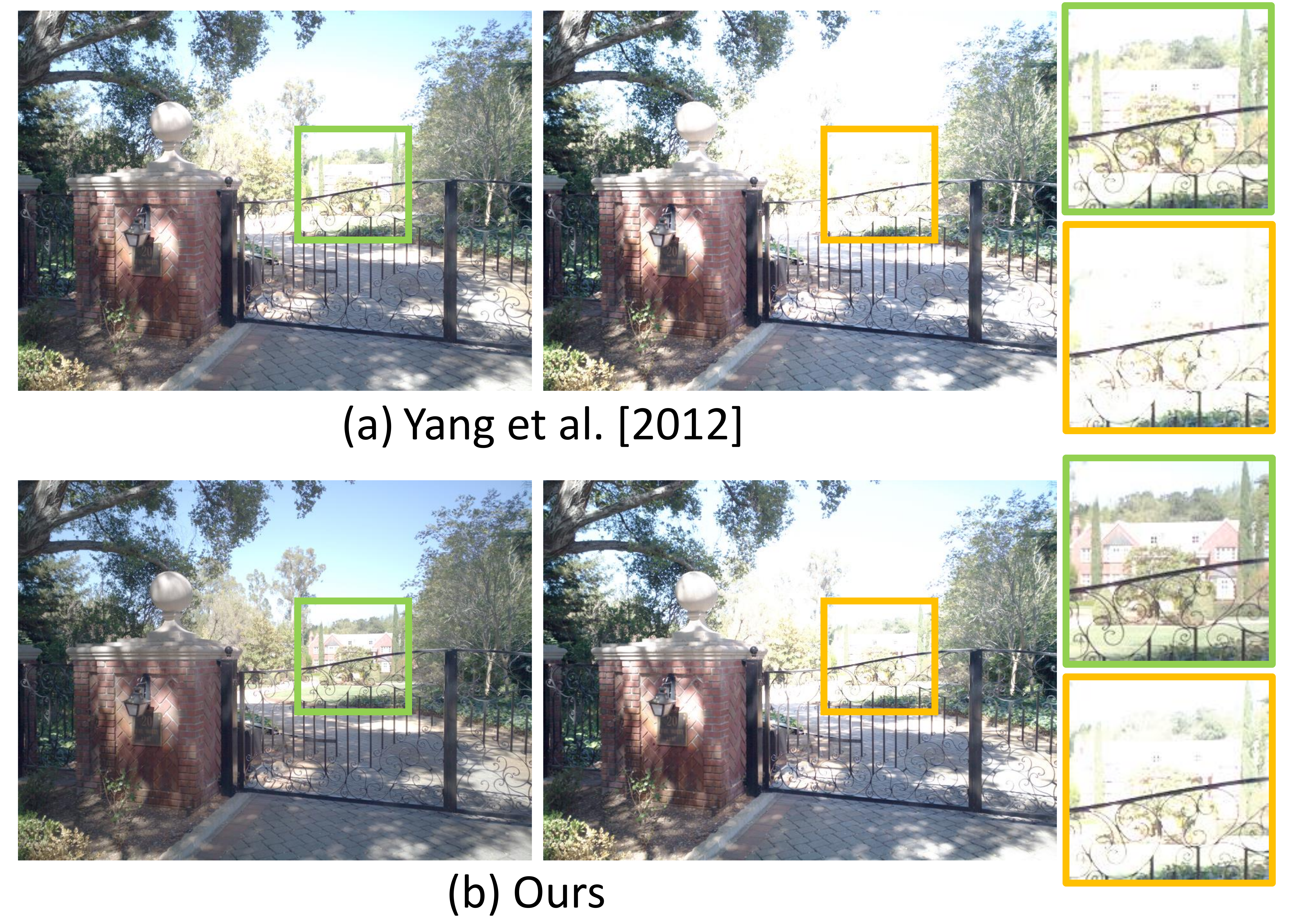}
\caption{Comparison with existing method~\cite{yang2012binocular}.}
\label{fig:yang_fail_case_over}
\end{figure}

Recently, {with the increased use of} stereoscopic devices, the notion of binocular tone mapping has been proposed by Yang et al.~\cite{yang2012binocular}. The key is to map an HDR image to two LDR images with different tone mapping parameters, one as left image and the other as right image, so that more human-perceivable visual content can be presented with the binocular LDR image pair (Fig.~\ref{fig:intro_pair_lrmean}(b)) than any single LDR image (Fig.~\ref{fig:intro_pair_lrmean}(a) or either image of Fig.~\ref{fig:intro_pair_lrmean}(b)) .  It is shown in the existing literature that when the left image and the right image are different in color and contrast, the fused vision by observing the binocular image pair via stereoscopic glasses is a non-linear combination, which may be able to preserve content from both views. Given an HDR image and any tone-mapped LDR image, Yang et al. proposed to look for another LDR image tone-mapped from the HDR input that has the largest visual difference with existing one, while preserving binocular fusibility for the LDR image pair, via an optimization framework. However, their method has three major problems. First, their method needs to fix one view and optimize the other, so it is unable to acquire an optimal binocular pair simultaneously from a given HDR image. Second, instead of directly measuring the amount of visual content in the generated binocular pair, they measure the visual difference between the two images and assume that large visual difference indicates more visual content, which may not be the case. An example is shown in Fig.~\ref{fig:yang_fail_case_over}(a). Both left and right images have high overall contrast and compressed local details, while the visual difference between left and right images is also quite large. However, the local details are not preserved in either view, and therefore is lost in the fused vision as well. Third, their method is based on the visible difference predictor (VDP) which is quite slow.

To resolve the above issues, in this paper, we proposed a novel {\em perception-based} binocular tone mapping method, that can generate an optimal binocular image pair (a left image and a right image) from an HDR image that presents the most visual content by designing a binocular perception metric. Comparing to Yang et al.'s method, our method has three major advantages. First, for any given HDR input, our method automatically obtains an optimal binocular image pair, i.e. the left and right images, simultaneously. Second, instead of measuring the visual difference between the left and right images, we propose a binocular perception metric that directly evaluates the amount of visual content of a binocular image pair, so that we can obtain the optimal binocular image pair with maximal visual content. Third, our method performs much more efficiently without using VDP.

Our binocular perception metric is designed based on the finding that overall contrast and local details are the two most important elements for enhancing visual content in binocular perception. The optimal binocular image pair that presents the maximal visual content should contain maximal overall contrast and maximal local details simultaneously, while preserving binocular fusibility. Since humans tend to preserve the vision from both left and right images in the fused vision, it is possible to encourage the left and right images to preserve overall contrast and local details in a separate manner. That is, we may encourage the left image to preserve local details as much as possible, and encourage the right image to preserve overall contrast as much as possible, while binocularly being fusible. In practice, we formulate the problem as an optimization where the energy function consists of three terms: contrast preservation term, detail preservation term, and fusibility term. By optimizing these three terms, we are able to obtain a binocular image that preserves maximal visual content while being fusible.

To validate our method, we apply it on HDR images of different genre and content. Qualitative and quantitative evaluations, as well as user study, are conducted. Statistics shows that our method outperforms the existing method in terms of both visual and time performance. Our contributions can be summarized as follows:
\begin{itemize}
	\item 	We make the first attempt to generate the optimal left and right image pair simultaneously from an input HDR image.
	\item	We propose a binocular perception metric for measuring the amount of visual content based on overall contrast and local details.
	\item 	Our method outperforms the existing method in terms of both visual and time performance.
\end{itemize}

\begin{figure*}[!ht]
	\centering
	\includegraphics[width=\linewidth]{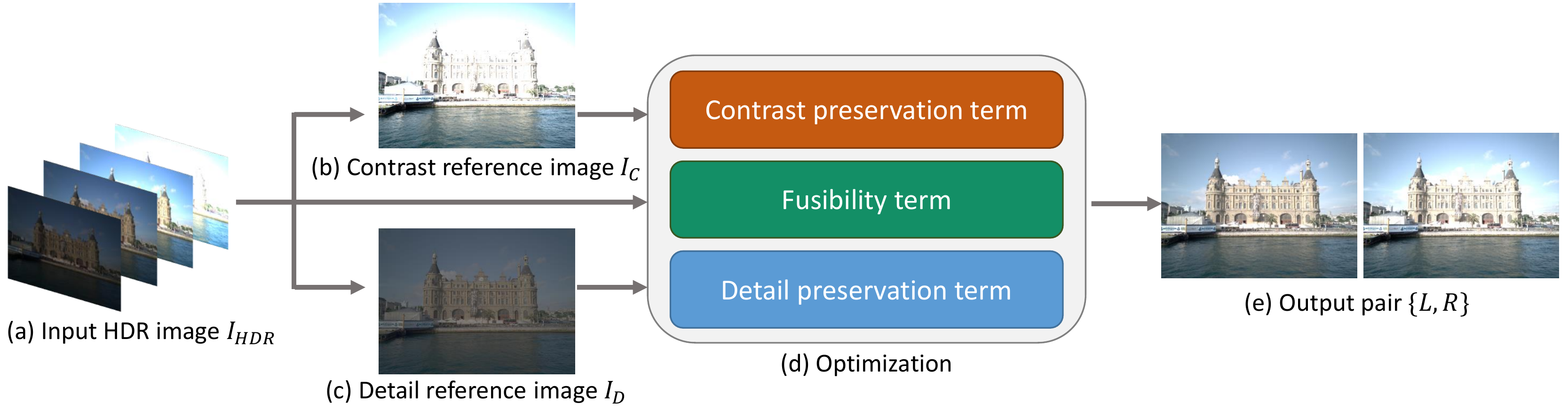}
	\caption{System overview.}
	\label{fig:overview_opt}
\end{figure*}

\section{Related Works}

The related works can be roughly classified into tone mapping operators, visible difference predictors, and binocular perception.


\subsection{Tone Mapping Operators}
Various tone mapping techniques have been proposed to map HDR images to LDR images. A comprehensive study has been conducted by Reinhard et al.~\cite{reinhard2005HDR}. In particular, Durand et al. proposed a bilateral tone mapping operator which decomposes an image into a base layer and a detail layer, where the base layer can be manipulated to change the overall contrast~\cite{durand2002fast}. Farbman et al. proposed to decompose and manipulate tone and detail based on the weighted least squares (WLS) framework in a multi-scale manner~\cite{farbman2008edge}.
To manipulate image detail in different scales without introducing artifacts, Local Laplacian Filter~\cite{paris2011local,aubry2014fast} is proposed that can work effectively and efficiently.

While the above tone mapping operators are designed to map an HDR image to a single LDR image, Yang et al. proposed a binocular mapping strategy by fixing one view and optimizing the other~\cite{yang2012binocular}. However, as we have shown in the previous section, Yang et al.'s method does not control the amount of visual content in the generated binocular pair, thus is unable to generate an optimal binocular image pair that preserves maximal visual content from an HDR image. In comparison, our method also generates a binocular LDR image pair from an HDR image, but via an optimization that simultaneously generates an optimal image pair that preserves maximal visual content. While our method is based on the bilateral tone mapping operator~\cite{durand2002fast}, we can replace it with other tone mapping operators, or even a combination of multiple operators to achieve better results, but with a trade-off of time performance due to the complexity of the other tone mapping operators.

\subsection{Visible Difference Predictors}
Predicting the visibility of distortions is a challenging problem.
Various visibility metrics that focus on modeling spatial contrast sensitivity have been proposed~\cite{zhang1997spatial, lubin1995visual, daly1992visible, mantiuk2011hdr}.
In particular, Zhang and Wandell proposed the sCIELab metric which utilizes a spatio-chromatic contrast sensitivity function prior to prefilter CIELab encoded pixels in order to compute the visibility map~\cite{zhang1997spatial}. 
Visual Discrimination Model (VDM)~\cite{lubin1995visual}, Visible Differences Predictor (VDP)~\cite{daly1992visible}, and HDR-VDP~\cite{mantiuk2011hdr} are tailored to predict visible difference in more complex images.
The major factors considered in these researches include luminance adaptation, contrast sensitivity, contrast masking, and frequency-selective visual channels~\cite{chandler2013seven}.
Recently, Butteraugli has been proposed to estimate the psychovisual similarity of two images and is adopted as the core part of Google's perceptually guided JPEG encoder 'Guetzli'~\cite{alakuijala2017guetzli}.
It accesses the visibility of artifacts based on feature maps instead of low-level contrast.

With a similar goal of our paper, Yang et al.~\cite{yang2012binocular} proposed to generate a pair of images that preserves more visual content by maximizing the visible difference between the generated pair. However, as we have stated, a binocular image pair of large visible difference may not preserve maximal visual content, as illustrated in Fig.~\ref{fig:yang_fail_case_over}.
In sharp comparison, we propose a binocular perception strategy to evaluate the amount of visual content presented in a binocular image pair. Based on the proposed binocular perception strategy, our method obtains the optimal image pair that presents the maximal visual content for any given HDR image.

\subsection{Binocular Perception}
Various psychological studies have explored the binocular perception of an image pair, where brightness, contour, and contrast are mostly studied.
Binocular brightness can be estimated with the local brightness amplitude in the two views.
Grossberg and Kelly conducted a survey on brightness perception~\cite{grossberg1999neural}, and studied a series of parameterized brightness models~\cite{levelt1965binocular, engel1969autocorrelation, de1974binocular, curtis1978binocular, legge1984binocular}.
These studies suggest that the binocular brightness of an image pair is approximately a linear combination, as long as the two views are not too different with each other. Therefore, the overall contrast of a binocular pair is approximately the averaged overall contrast of the two views.
Different from the brightness perception, contrast perception suggests that the contrast (local details) in one monocular view generally inhibits the other~\cite{wilson2017binocular}. This shows that the fused vision of a binocular image pair tends to preserve the local details from both images.
Meese et al. conducted a survey on contrast perception~\cite{meese2006binocular}, and studied a series of parameterized contrast models~\cite{legge1981binocular, meese2004low, meese2005interocular, maehara2005binocular}.
Besides, Chua et al.~\cite{hau2015colorbless} explored the ability of binocular luster effect, the salient shininess of some parts of fused vision, for enhancing visual information.
Binocular fusibility is also explored in the existing research~\cite{yang2012binocular} based on a series of psychological studies~\cite{levelt1965binocular, liu1992failure, steinman2000foundations}. A binocular visual comfort predictor (BVCP) metric is proposed to measure the fusibility of a binocular image pair by Yang et al.~\cite{yang2012binocular}. Later researches followed BVCP on evaluating binocular fusibility~\cite{shen2016seamless}.
We also follow BVCP in evaluating the binocular fusibility of an image pair, but we change the original binary classification to a soft predictor that predicts a fusibility value for each pixel, while remaining faithful to the original threshold values, to fit into our optimization framework.

\section{Overview}

Our system is overviewed in Fig.~\ref{fig:overview_opt}. To generate a LDR image from the input HDR image, we adopt the bilateral tone mapping operator proposed by Durand et al.~\cite{durand2002fast}. This bilateral tone mapping operator decomposes the output LDR image into a base layer which mainly captures the image contrast and a detail layer which mainly captures the details. By adjusting the contrast of the base layer, we can manipulate the generated LDR image. In our framework, given an input HDR image $I_{HDR}$ (Fig.~\ref{fig:overview_opt}(a)), we intend to generate a binocular LDR image pair $\{L,R\}$, consisting of a left image $L$ and a right image $R$, that presents the maximal visual content while remaining binocularly fusible (Fig.~\ref{fig:overview_opt}(e)). Both the left image $L$ and the right image $R$ are tone mapped from the input HDR image $I_{HDR}$, with different tone mapping operators as
\begin{eqnarray}
L = \mathbf{T}(I_{HDR}, \beta_L)\\
R = \mathbf{T}(I_{HDR}, \beta_R)
\end{eqnarray}
where $\mathbf{T}(I_{HDR}, \beta)$ is the bilateral tone mapping operator that maps an HDR image $I_{HDR}$ to a LDR image. $\beta_L$/$\beta_R$ are the parameters used in the bilateral tone mapping operator, and denote the contrast of the decomposed based layer in the left/right image respectively. To optimize the generated binocular image pair $\{L,R\}$, we are actually optimizing the corresponding tone mapping parameters $\beta_L$ and $\beta_R$. Note that we fix the other parameters used in the bilateral tone mapping operator to the suggested default values since they have very little influence on the generated results.

Note that while our method is based on the bilateral tone mapping operator, we can actually replace it with other tone mapping operators, as long as the operators have parameters that can effectively control overall contrast and local details. It is also possible to combine multiple operators together. While the visual quality might be improved, the time complexity will also be significantly increased. During optimization, to avoid trapping in local optimum, we first optimize overall contrast only and adopt the optimized result as the initial point. Then we optimize both the overall contrast and the local details simultaneously. This strategy is quite straightforward but effective.

In order that the generated binocular image pair preserve the most visual content, we mean to maximize the perceived overall contrast and local details, while preserving binocular fusibility. In practice, we formulate an optimization problem by minimizing an energy function that contains three corresponding energy terms: contrast preservation term, detail preservation term, and fusibility term (Fig.~\ref{fig:overview_opt}(d)). The contrast preservation term and the detail preservation term are used to ensure that the generated binocular image pair preserves enough overall contrast and local details respectively. The fusibility term is used to ensure that the generated left and right images does not cause visual discomfort. To evaluate how much overall contrast and local details are preserved in the generated binocular image pair, we propose to compare the approximated binocular fusion with two reference images, an contrast reference image $I_C$ that captures the maximal overall contrast (Fig.~\ref{fig:overview_opt}(b)) and a detail reference image $I_D$ that captures the maximal local details (Fig.~\ref{fig:overview_opt}(c)). The details of our  optimization is presented in Section~\ref{sec:approach}.

\section{Optimization}
\label{sec:approach}

We formulate an optimization problem by minimizing an energy function that contains three energy terms: contrast preservation term, detail preservation term, and fusibility term. To calculate the contrast preservation term and the detail preservation term, we propose generating two reference images, an contrast reference image and a detail reference image, so that the overall contrast and the local details of the output image pair can be measured based on the reference images.

In this section, we first show why and how we generate the reference images, and then present how we formulate the three terms.

\subsection{Reference Image Generation}
To evaluate the how much overall contrast and local details are preserved in the generated binocular image pair, we need to compare the approximated binocular fusion with the input HDR image in terms of both overall contrast and local details. However, there lack direct measurements for evaluating overall contrast and local details in HDR images. Therefore, we propose to represent the original HDR image by two reference images, a contrast reference image that captures the maximal overall contrast, and a detail reference image that captures the maximal local details. Then we compare the approximated binocular fusion with these two reference images respectively to measure how much overall contrast and local details are preserved in the generated binocular image pair comparing to the original HDR image. 

The two reference images are also tone mapped from the input HDR image $I_{HDR}$ via the bilateral tone mapping operator~\cite{durand2002fast}. In particular, to generate the contrast reference image that captures most overall contrast of the HDR image, we set the contrast of the decomposed basic layer to a relatively high value. On the other hand, to generate the detail reference image that captures most local details, we set the contrast of the decomposed basic layer to a relatively low value. In practice, we set it to $6.0$ and $1.5$ for the contrast reference image and the detail reference image respectively. Thus, we formulate the contrast image $I_C$ and the detail reference image $I_D$ as
\begin{eqnarray}
I_C = \mathbf{T}(I_{HDR}, 6.0)\\
I_D = \mathbf{T}(I_{HDR}, 1.5)
\end{eqnarray}

\begin{figure}[!ht]
	\centering
	\includegraphics[width=\linewidth]{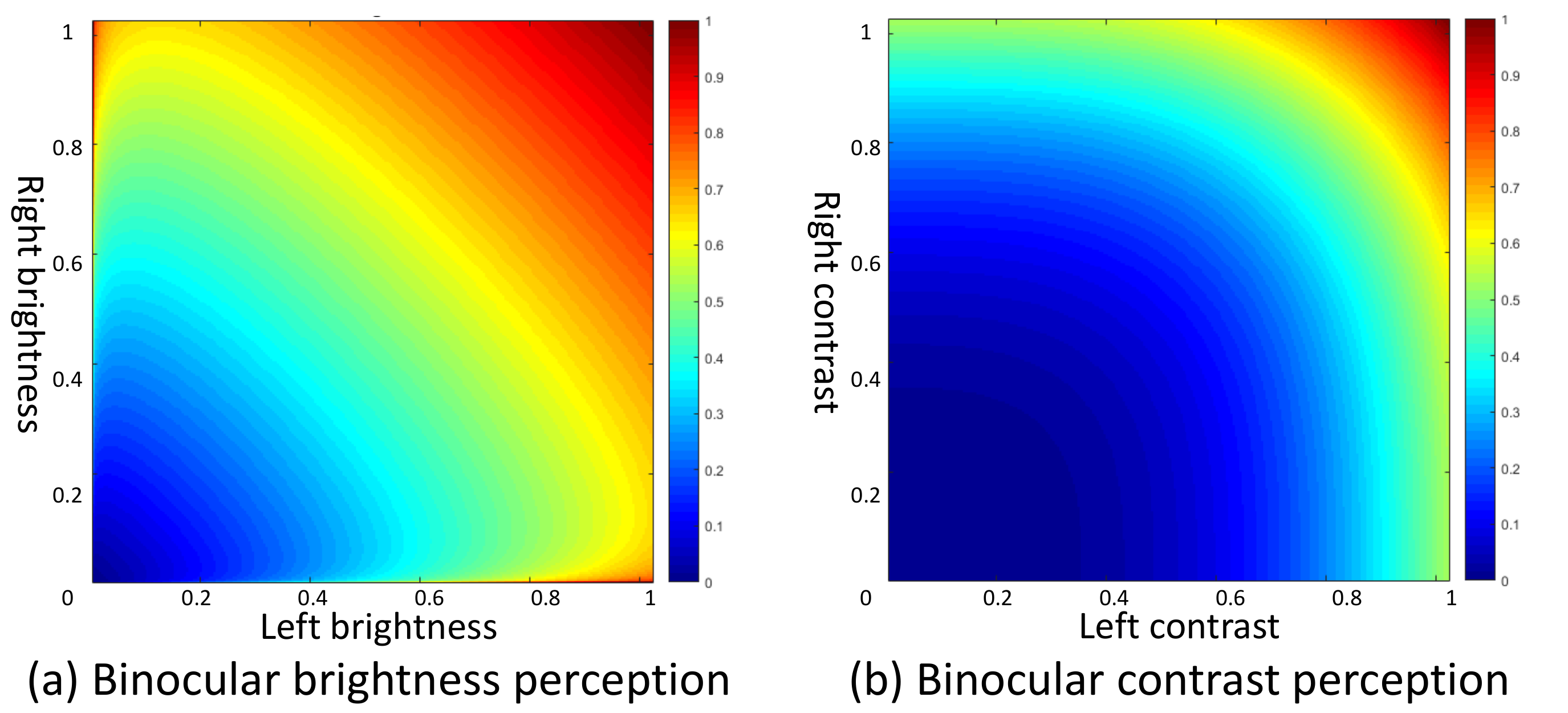}
	\caption{Plot of binocular perception.}
	\label{fig:bino_isofigure}
\end{figure}

\subsection{Contrast Preservation Term}

To measure how a binocular LDR image pair $\{L,R\}$ preserves the overall contrast of an HDR image $I_{HDR}$, we actually measure how the binocular LDR image pair preserves the brightness of the contrast reference image $I_C$ over all pixels. Since the contrast reference image $I_C$ already captures most overall contrast of the input HDR image, the output LDR image pair would also preserve the overall contrast as long as the fused pixels has similar brightness values with $I_C$. The existing brightness perception studies show that the binocular brightness of an image pair is approximately a linear combination of the two images, as long as the two views are not too different with each other~\cite{levelt1965binocular, engel1969autocorrelation, de1974binocular, curtis1978binocular, legge1984binocular}. In particular, we adopt the the model proposed by Curtis and Rule~\cite{curtis1978binocular}. Given a binocular LDR image pair $\{L,R\}$ and a pixel $p$, the fused binocular brightness value of $p$ can be approximated via the equation
\begin{eqnarray}
b_{L,R}(p) = \sqrt{b_L(p)^2+b_R(p)^2+2b_L(p)b_R(p)\cos\alpha}
\label{eqn:bino_brightness}
\end{eqnarray}
where $b_L(p)$ and $b_R(p)$ are the average brightness value of $p$ in the left image $L$ and the right image $R$ in a local neighborhood respectively. The reason that we use the average brightness in a local neighborhood is because that the binocular vision is fused within a fusion area instead of a single pixel~\cite{yang2012binocular}. While the fusion area is related to the DPI of the monitor, we adopt the same setting with Yang et al.'s work, where the radius of the fusion area is about $16$ pixels. $\alpha$ is fixed to $120^\circ$ as suggested by Grossberg et al.~\cite{grossberg1999neural}. We visualize this binocular brightness perception in Fig.~\ref{fig:bino_isofigure}(a). We can observe that the combination is approximately linear. Then we can define the contrast preservation term as
\begin{eqnarray}
E_c(L,R) =\frac{1}{N} \sum_{p} \frac{|b_{I_C,I_C}(p) - b_{L,R}(p)|}{|b_{I_C,I_C}(p) - b_{I_D,I_D}(p)|+\epsilon}
\end{eqnarray}
where $N$ is the total number of pixels. $\epsilon$ is a very small number to avoid dividing by zero. The reason that we adopt $b_{I_C,I_C}(p)$ as the reference brightness value is to reduce the influence introduced by the difference of monocular view and binocular fusion. $|b_{I_D,I_D}(p)-b_{I_C,I_C}(p)|$ is a normalization factor. The more contrast the binocular pair is preserved, the smaller the contrast preservation term $E_c(L,R)$ will be. If the binocular pair preserves full overall contrast, this term will be minimized to 0.

\subsection{Detail Preservation Term}
To measure how a binocular LDR image pair $\{L,R\}$ preserves the local details of an HDR image $I_{HDR}$, we actually measure how the binocular LDR image pair preserves the local details of the detail reference image $I_D$ over all pixels. The preservation of local details can be evaluated by measuring the difference of local contour contrast. Here, the local contour contrast of a pixel $p$ is defined as the difference between the maximal and minimal brightness values within a local neighborhood
\begin{equation}
	c(p) = \max_{\Omega(p)}(b(p)) - \min_{\Omega(p)}(b(p))
\end{equation}
where $b(p)$ is the brightness value of $p$, and $\Omega(p)$ denotes a $3\times3$ local neighborhood of $p$. The existing studies show that the binocular perception of local contrast is non-linear, where the view with large local contrast generally inhibits the other view~\cite{wilson2017binocular,legge1981binocular, meese2004low, meese2005interocular, maehara2005binocular}. In particular, we adopt the Legge-type model~\cite{legge1981binocular}. Given a binocular LDR image pair $\{L,R\}$ and a pixel $p$, the fused binocular local contrast of $p$ can be approximated via the equation
\begin{eqnarray}
c_{L,R}(p) = \dfrac{(c_L(p)^s+c_R(p)^t)^{s/t}}{z+c_L(p)^s+c_R(p)^t}
\label{eqn:bino_contrast}
\end{eqnarray}
where $c_L(p)$ and $c_R(p)$ and the local contrast of $p$ in the left image $L$ and the right image $R$ respectively. $s = 3.47$, $t = 3.03$ and $z = 4.76$ are fixed parameters as suggested by Meese et al.\cite{meese2006binocular}. We visualize this binocular local contrast perception in Fig.~\ref{fig:bino_isofigure}(b). We can observe that the combination is non-linear. Then we can define the detail preservation term as
\begin{eqnarray}
E_d(L,R) = \frac{1}{|\mathbb{E}|}\sum_{p\in \mathbb{E}} \frac{\mathbf{H}(c_{I_D,I_D}(p)-c_{L,R}(p))}{\mathbf{H}(c_{I_D,I_D}(p) - c_{I_C,I_C}(p))+\epsilon}
\end{eqnarray}
where
\begin{equation}
   \mathbf{H}(x)=
\begin{cases}
0 &\text{if $x<0$} \\
x &\text{if $x\geq 0$}.
\end{cases}
\end{equation}
$\mathbb{E}$ is the set of edge pixels in $I_D$. Here, we only consider the edge pixels in $I_D$ to focus on the pixels exhibiting local details and avoid the dominance of non-edge pixels. To detect the edge pixels, we adopt Canny edge detector ~\cite{canny1986computational} with default parameters, which was also used by Yang et al.~\cite{yang2012binocular} and Shen et al.~\cite{shen2016seamless}. We adopt the piecewise function $\mathbf{H}(x)$ to measure the difference of local contrast in order that we only penalize pixels where $I_D$ contains local details but lost in the generated binocular image pair, but not conversely. Similar with the contrast preservation term, $\mathbf{H}(c_{I_D,I_D}(p) - c_{I_C,I_C}(p))$ is a normalization factor. The more details the binocular image pair preserves, the smaller the detail preservation term $E_d(L,R)$ will be. If the binocular pair preserves full local details, this term will be minimized to 0.

\subsection{Fusibility Term}

Besides preserving the overall contrast and local details of the HDR image, we also require that the generated LDR image pair is binocularly fusible. To measure the fusibility of two image, we adopt the binocular visual comfort predictor (BVCP) proposed by Yang et al.~\cite{yang2012binocular}, consisting of a contour fusion metric $B_{cf}$ and a region contrast fusion $B_{rf}$. If either $B_{cf} > 1$ or $B_{rf} > 1$, it means that the binocular image pair is not fusible, and therefore causing visual discomfort.
To fit BVCP to our optimization framework, we formulate a continuous energy term based on $B_{rc}$ and $B_{cf}$ as
\begin{eqnarray}
E_f(L,R) = \phi(B_{rc}) + \phi(B_{cf})
\end{eqnarray}
where $\phi(x)= max(x - 1, 0)$ is an activation function.

We adopted the suggested parameters of the BVCP metric for measuring fusibility. It generates stable fusible pairs for most of the cases. But a small number of users still perceive unstable fusion for some of the cases. For these particular cases, we can use tighter fusibility parameters, but with a trade-off of preserving less visual content.

\subsection{Overall Energy Function}

By combining the above three energy terms, the overall energy function can be written as
\begin{eqnarray}
E(L,R) = E_c(L,R) + \lambda_1 E_d(L,R) + \lambda_2 E_f(L,R)
\label{eqn:energy_all}
\end{eqnarray}
where $\lambda_1$ and $\lambda_2$ are the weighting factors. 
If $\lambda_1$ is set to a relatively large value, it means that our optimization focuses more on local details than overall contrast, and vice versa.
To guarantee that the binocular fusibility of the image pair is satisfied in most pixels, we set $\lambda_2$ to a relatively high value. In practice, we set $\lambda_2=10$ which is large enough to meet the fusibility requirement. $\lambda_1$ is set to $1.25$ in all our experiments, which is decided via a user study (Section~\ref{sec:userstudy}). 

Recall that the left image $L$ and the right image $R$ are actually tone mapped from the HDR image $I_{HDR}$ with the tone mapping operator $\mathbf{T}$ and tone mapping parameters $\beta_L$ and $\beta_R$. The optimization function can actually be written as
\begin{eqnarray}
\{\beta_L^*, \beta_R^*\} = \argmin_{\{\beta_L, \beta_R\}} \ \  E(\mathbf{T}(I_{HDR}, \beta_L), \mathbf{T}(I_{HDR}, \beta_R))
\end{eqnarray}
The optimization can be solved with the gradient descent solver which generally converges in $30$ iterations.

\section{Results and Discussion}

\begin{figure*}[!htbp]
	\centering
	\includegraphics[width=.8\linewidth]{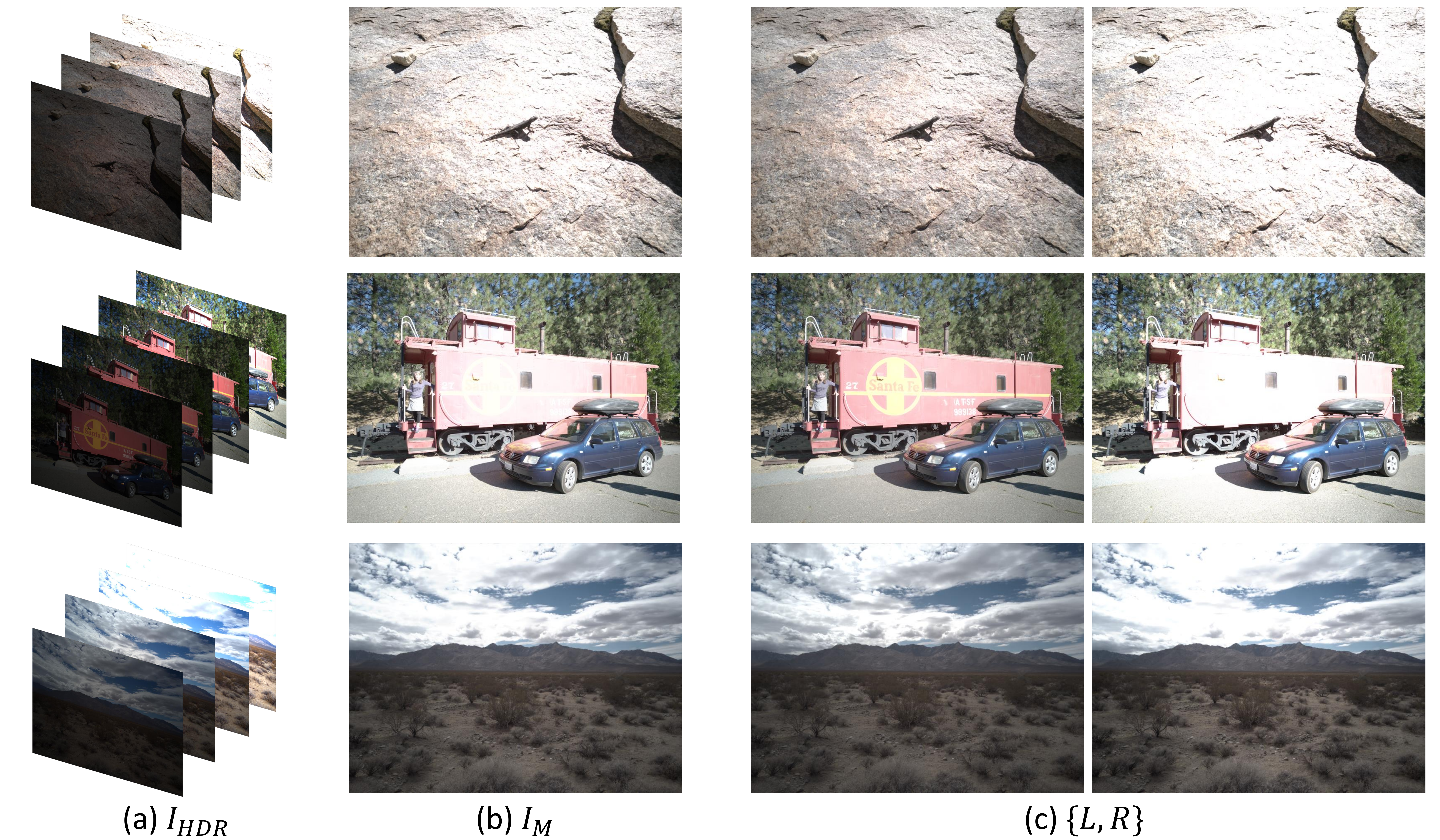}
	\caption{Comparison with monocular LDR image $I_M$.}
	\label{fig:result_cmp_lrmean}
\end{figure*}

\begin{figure*}[!htbp]
	\centering
	\includegraphics[width=\linewidth]{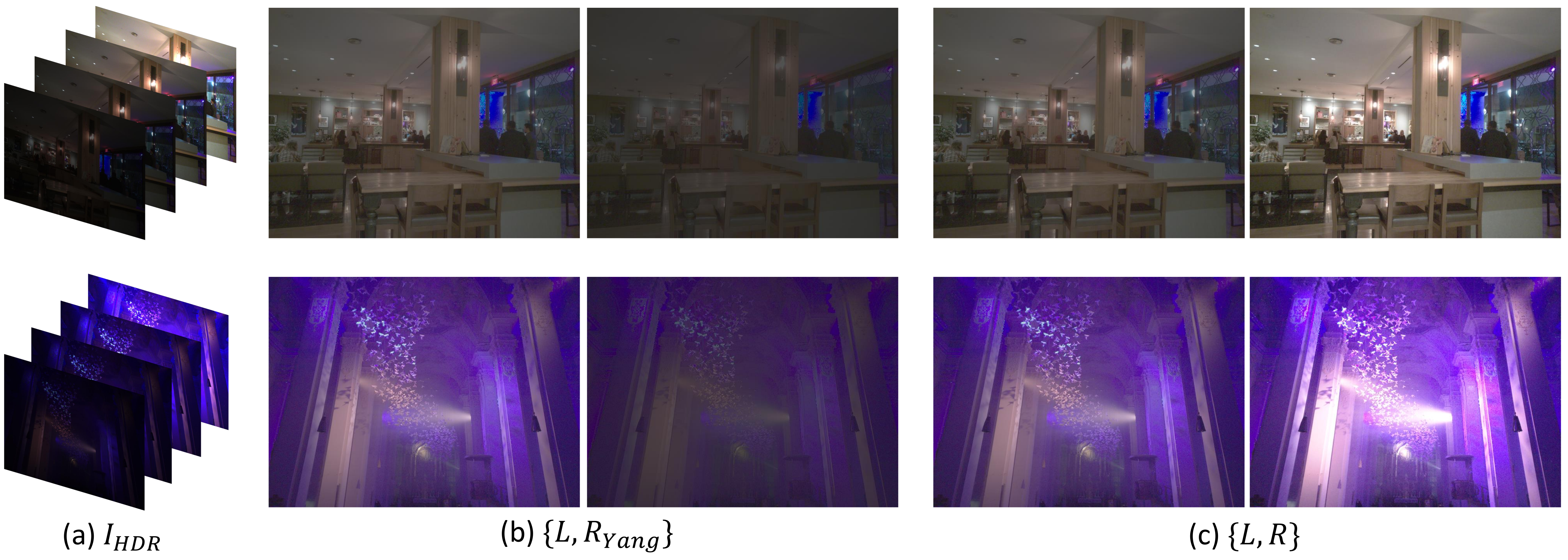}
	\caption{Comparison with Yang et al.~\cite{yang2012binocular} with left images (more details and lower contrast) fixed.}
	\label{fig:result_cmp_yang_detail}
\end{figure*}

\begin{figure*}[!htbp]
	\centering
	\includegraphics[width=\linewidth]{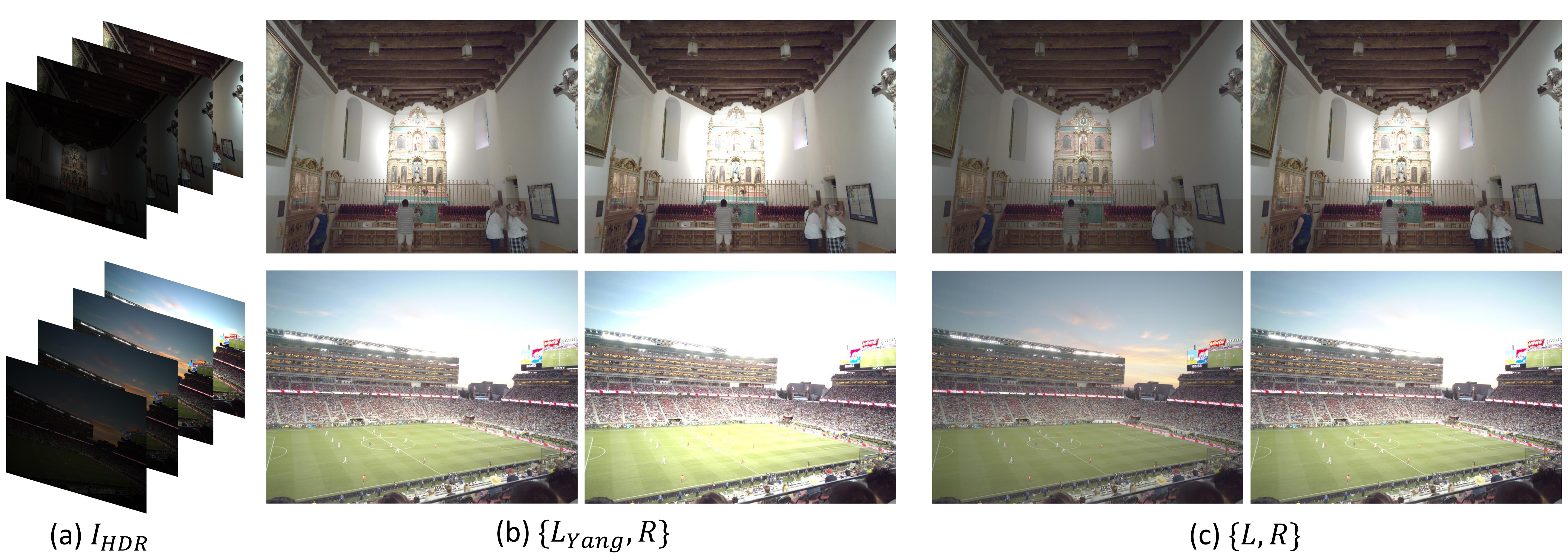}
	\caption{Comparison with Yang et al.~\cite{yang2012binocular} with right images (higher contrast and less details) fixed.}
	\label{fig:result_cmp_yang_contrast}
\end{figure*}

\begin{figure*}[!htbp]
	\centering
	\includegraphics[width=\linewidth]{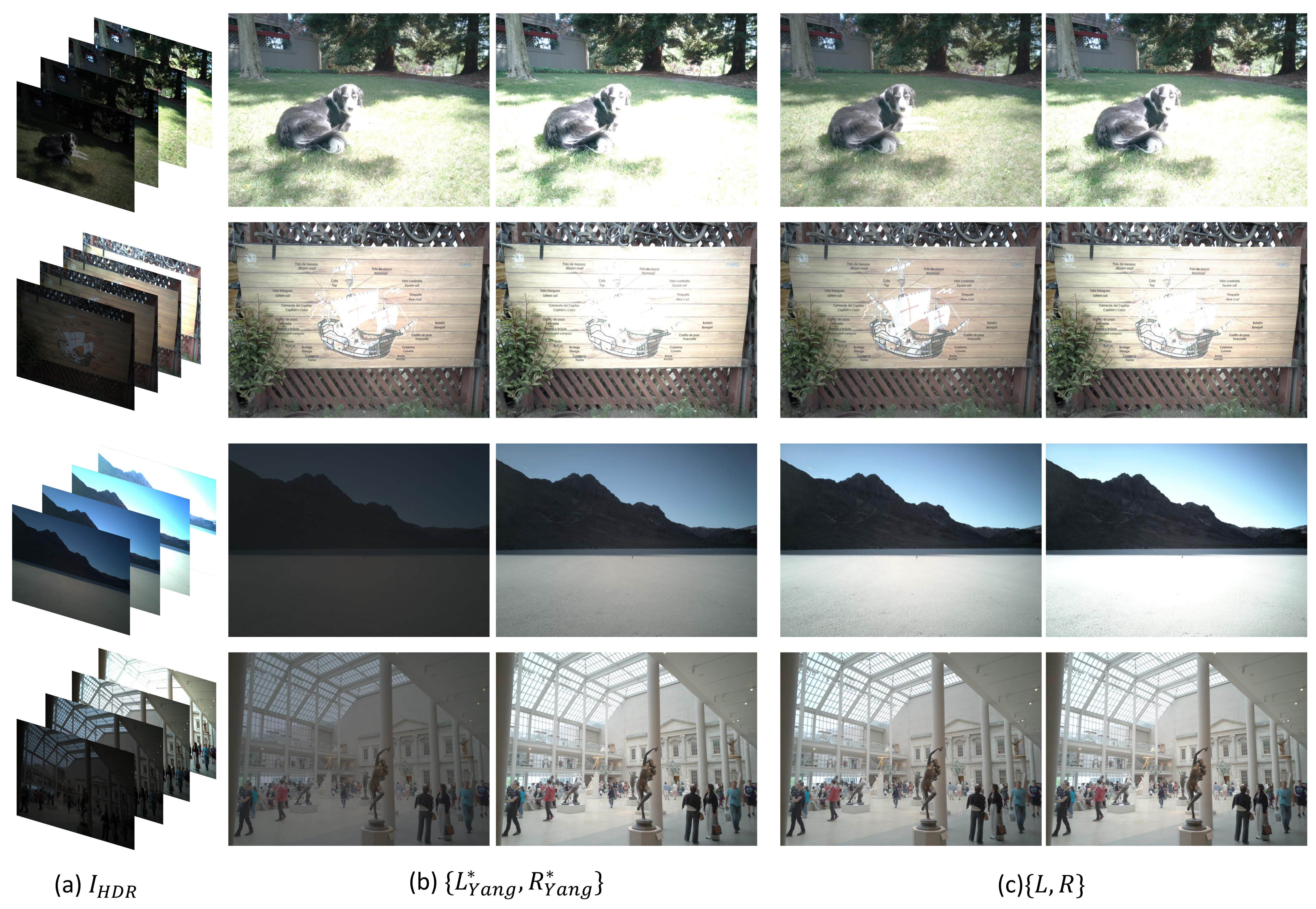}
	\caption{Comparison with Yang et al.~\cite{yang2012binocular} with brute-forced optimum.}
	\label{fig:result_cmp_yang_global}
\end{figure*}

\subsection{Qualitative Evaluation}

We evaluate our method on thousands of images of various genre and content, convincing results are obtained in all cases. Since our output is a binocular image pair which should be perceived via stereoscopic devices, we have put the original left and right images of all our results in the supplementary materials. Readers are encouraged to view the original image pairs via a stereoscopic display to see how our output image pair captures the overall contrast and the local details of the input HDR image simultaneously. 

\noindent\textbf{Monocular LDR vs. Binocular LDR Pair} Fig.~\ref{fig:result_cmp_lrmean} compares our binocular LDR image pair (Fig.~\ref{fig:result_cmp_lrmean}(c)) with a monocular LDR image (Fig.~\ref{fig:result_cmp_lrmean}(b)). For our binocular image pair, since we do not specifically require either view having higher overall contrast, either the left image or the right image may have higher overall contrast, and therefore less local details. As suggested by Yang et al.~\cite{yang2012binocular}, the left and right images can be swapped which leads to no influence on the fused vision. Therefore, for consistent visualization, we always present the view with more details (lower contrast) as the left image, and the view with higher contrast (less details) as the right image. For the monocular LDR image, we adopt the average of left and right image in the parameter space. Concretely speaking, while our left image $L$ and right image $R$ are tone mapped from the original HDR image $I_{HDR}$ with parameter $\beta_L$ and $\beta_R$ respectively, we generate the monocular LDR image with $I_M=\mathbf{T}(I_{HDR}, (\beta_L+\beta_R)/2)$ so that this LDR image shares similar overall contrast with the linear fusion of our binocular image pair. It is observable that our binocular image pair preserves more local details than the monocular LDR image, especially in high-brightness regions. 


\noindent\textbf{Comparison with Existing Method} Since the existing tone mapping method proposed by Yang et al.~\cite{yang2012binocular} requires to fix one view and optimize the other, we conduct two experiments to compare our results with theirs. In the first experiment, we use our left image $L$ (more details and lower contrast) as the fixed view and optimize the right image $R_{Yang}$ with their method. With the given left images, Yang et al. achieves similar results with our method in most cases as long as the overall contrast in the optimized right images are enlarged. But there are also cases where the right images have very low brightness values, with both low contrast and low details, even though the visual difference between the left and right images calculated by VDP is large due to the large difference in brightness as shown in Fig.~\ref{fig:result_cmp_yang_detail}(b). In comparison, our output image pairs always contain high contrast and details simultaneously to present more visual content (Fig.~\ref{fig:result_cmp_yang_detail}(c)). In the second experiment, we use our right image $R$ (higher contrast and less details) as the fixed view and optimize the left image $L_{Yang}$ with their method. While the fixed right images already have relatively high contrast, the optimized left images generated by Yang et al.'s method may have lower or even higher overall contrast as shown in Fig.~\ref{fig:result_cmp_yang_contrast}(b). In either case, the VDP between left and right images is still large. However, if the optimized left images have even higher contrast, the presented visual content only has small increment in overall contrast, and no increment in local details at all. In comparison, our binocular image pair presents more visual content (Fig.~\ref{fig:result_cmp_yang_contrast}(c)).

We further implemented an optimal solution for Yang et al.'s solution by searching in brute-force, more concretely, we quantize the parameter space and test all potential parameters for the generated left and right image. Then we find the optimal image pair $\{L_{Yang}^*, R_{Yang}^*\}$ that has largest visual difference based on VDP while being binocularly fusible based on BVCP. Fig.~\ref{fig:result_cmp_yang_global} shows the result. We can see that, while the VDP between left and right images are maximized, the two images may both have very low or very high brightness (Fig.~\ref{fig:result_cmp_yang_global}(b)). In either case, both the overall contrast and local details are limited, but still leading to large visual difference based on VDP. In sharp contrast, our method obtains binocular image pairs with proper contrast and details, and preserves more visual content for the input HDR image (Fig.~\ref{fig:result_cmp_yang_global}(c)).


\begin{figure*}[!htbp]
	\centering
	\includegraphics[width=\linewidth]{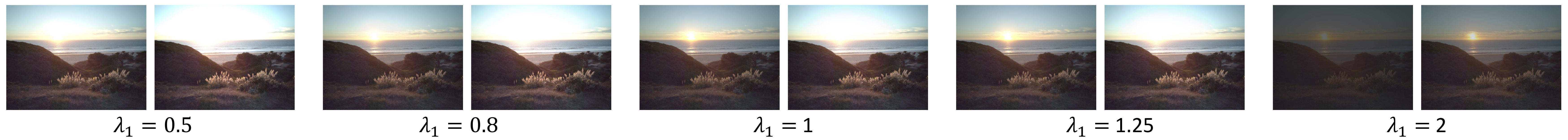}
	\caption{Results with different $\lambda_1$.}
	\label{fig:result_cmp_lambda1}
\end{figure*}
 
\subsection{Quantitative Evaluation}

We further evaluate our system quantitatively on the HDR+ Burst Photography Dataset~\cite{hasinoff2016burst} composed of 3,620 natural images. We adopt both our proposed binocular perception metric ($E_c$, $E_d$, and $E$) and the VDP used in~\cite{yang2012binocular}. Similar with the qualitative evaluation, we also compare our method with the monocular LDR output $I_M$ and three variants of Yang et al.'s method, i.e. $\{L, R_{Yang}\}$, $\{L_{Yang}, R\}$, and $\{L_{Yang}^*, R_{Yang}^*\}$, as defined in the previous subsection. To avoid the difference between monocular and binocular vision, the monocular LDR output is evaluated as a binocular image pair $\{I_M, I_M\}$. We also compare with the two reference images, the contrast reference image $I_C$ and the detail reference image $I_D$, quantitatively. Similar with $I_M$, $I_C$ and $I_D$ are also evaluated as binocular image pairs $\{I_C, I_C\}$ and $\{I_D, I_D\}$.

The statistics is shown in Table~\ref{table:energy_val}. Note that we do not show binocular fusibility in our statistics because all methods achieve binocular fusibility in almost all pixels. Intuitively, a smaller $E_c$ indicates better overall contrast preservation. A smaller $E_d$ indicates better local detail preservation. A smaller $E$ indicates better visual content preservation. A larger VDP value indicates larger visual difference in the output image pair. We can observe that $\{L_{Yang}^*, R_{Yang}^*\}$ achieves the highest score for VDP since they are attempting to maximize the VDP of the output image pair. It also achieves the highest score for $E_c$ since the VDP difference is mainly presented as brightness difference, but with a trade-off for the local details $E_d$, and therefore the overall visual content $E$.  $\{L_{Yang}, R\}$ achieves a low score for $E_c$ since it can preserve the contrast, but with a trade-off for local details $E_d$. The scores of both $\{L_{Yang}^*, R_{Yang}^*\}$ and  $\{L_{Yang}, R\}$ show that maximal visual difference does not necessarily indicate maximal visual content. This also conforms to the qualitative evaluation. $\{L, R_{Yang}\}$ and our method achieve similar scores for all metrics. This is because that we have already optimized the local details in the fixed left image, and their method only needs to find an optimal right image with large overall contrast, which is exactly what VDP is focusing on.

\begin{table}[!htbp]
	\centering
	\begin{tabular}{c||ccc|c}
	\hline
	Score 	& $E_c$ & $E_d$ & $E$ & VDP   		\\ 
	\hline
	
	$\{L, R\}$							&0.2698 			&0.3018 			&\textbf{0.5716} 	&0.0286 	 \\
	$\{I_M, I_M\}$ 						&0.2684 			&0.4133 			&0.6817 			&0 	 \\ 
	$\{L, R_{Yang}\}$					&0.2793 			&\textbf{0.2986} 	&0.5779 			&0.0294 	  \\
	$\{L_{Yang}, R\}$					&0.2063		 		&0.3704 			&0.5767 			&0.0347   \\ 
	$\{L_{Yang}^*, R_{Yang}^*\}$		&\textbf{0.1960} 	&0.4081 			&0.6041 			&\textbf{0.0443} 	 \\ 
	\hline
	$\{I_D, I_D\}$ 						& 1 	& 0	  & 1 	& 0 \\
	$\{I_C, I_C\}$ 						& 0     & 1   & 1.25 	& 0 \\
	\hline
\end{tabular}
	\caption{Statistics of quantitative evaluation.}
	\label{table:energy_val}
\end{table}

\subsection{User study}
\label{sec:userstudy}

We conducted two user studies: one to learn the optimal parameter $\lambda_1$ in Eq.~\ref{eqn:energy_all}, the other to compare our method with other methods in terms of human perception. We invite 12 participants for both user study, including 8 males and 4 females, aging from 20 to 35. $20$ test images are involved in the experiments. We show the images on a ASUS G750JX laptop with 3D display. The displaying luminance is 250 \text{$cd/m^2$}. Users are asked to sit half meter away from the display. Throughout the test, participants are asked to wear 3D glasses.

\noindent\textbf{Parameter Evaluation} To obtain the optimal parameter $\lambda_1$ in Eq.~\ref{eqn:energy_all}. For each test image, we generate 5 output image pairs with $\lambda_1$ set to 0.5, 0.8, 1, 1.25, and 2 respectively (Fig.~\ref{fig:result_cmp_lambda1}) and present them to the user side by side. Note that the positions of the five results are randomly placed. Then we ask the user to select one from the five images that they like more in terms of overall visual experience. The statistics is shown in Table~\ref{table:user_study_lambda}. As a result, $\lambda_1=1.25$ is most preferred by the users. This suggests that users generally focus more on local details than overall contrast. We fix $\lambda_1$ to $1.25$ in all our experiments.

\begin{table}[!htbp]
	\centering
	\begin{tabular}{c|ccccc}
		\hline
		$\lambda_1$&0.5 & 0.8 & 1.0 & 1.25 & 2.0 \\ \hline
		Preference & 0.04 &  0.27  &  0.30   &  0.32   &  0.07 \\ \hline
	\end{tabular}
	
	\caption{User study for $\lambda_1$ analysis.}
	\label{table:user_study_lambda}
\end{table}

\noindent\textbf{Comparison with Existing Method} Similar with qualitative and quantitative evaluations, we also compare our method with the monocular LDR output $I_M$ and three variants of Yang et al.'s method, i.e. $\{L, R_{Yang}\}$, $\{L_{Yang}, R\}$, and $\{L_{Yang}^*, R_{Yang}^*\}$, as defined in the previous subsections. To avoid the visual difference between monocular and binocular vision, $I_M$ is also presented as a binocular image pair $\{I_M, I_M\}$. Since it is difficult for users to directly rate each binocular image pair, we present two binocular image pairs to the users side by side, where one is our result and the other is the result of an existing method. Note that the positions of the two images are randomly swapped to avoid bias. Then we ask the user to select one from the two images that they like more in terms of overall visual experience, one that presents better overall contrast, and one that preserves more local details. If the user selects our result, we mark the score as 1. Otherwise, we mark it as 0. Then we average the scores for all testing cases and the statistics is shown in Table~\ref{table:user_study}. Intuitively, if the score is larger than 0.5, it means that the user prefers our results than the other. The larger the score is, the more preferred our results are. We can observe that our method and $\{L,R_{Yang}\}$ achieve similar preference because that we have already optimized the local details in the fixed left image, and their method (the VDP) works well in find an optimal right image with large overall contrast. On the other hand, when compared with $\{I_M, I_M\}$, $\{L_{Yang}, R\}$ and $\{L_{Yang}^*, R_{Yang}^*\}$, our method is more preferred by the user in terms of all metrics.

\begin{table}[!htbp]
	\centering
		\begin{tabular}{cc|c|c|cc}
			\hline
			\multirow{2}{*}{}& 
			\multirow{2}{*}{}  &\multirow{2}{*}{Mean} & \multirow{2}{*}{Std.} &  \multicolumn{2}{c}{95\% Conf. Interval} \\ \cline{5-6}
			&& &	 & LWR Bnd. & UPR Bnd.\\ \hline
			\multirow{3}{*}{(a)}
			& Contrast	& 0.821 	&0.122 		&0.764 		&0.878 	\\
			& Detail  	& 0.783 	&0.099  	&0.737		&0.830 \\
			& Prefer. 	& 0.829 	&0.099 		&0.783  	&0.876 \\
			\hline

			\multirow{3}{*}{(b)}
			& Contrast	& 0.475 	&0.143 		&0.408		&0.542  \\
			& Detail  	& 0.529 	&0.154 		&0.457 		&0.601 \\
			& Prefer. 	& 0.513 	&0.156 		&0.440 		&0.585 \\
						\hline

			\multirow{3}{*}{(c)}
			& Contrast	& 0.688 	&0.104 		&0.639		&0.736 \\
			& Detail  	& 0.713 	&0.147 		&0.644 		&0.781  \\
			&Prefer. 	& 0.767 	&0.100 		&0.720 		&0.813 \\
			\hline
						\multirow{3}{*}{(d)}
			& Contrast	& 0.825 	&0.140 		&0.759 		&0.891 \\
			& Detail  	& 0.771 	&0.129 		&0.710		&0.831 \\
			& Prefer. 	& 0.883 	&0.091 		&0.841 		&0.926 \\
			
					\hline
		\end{tabular}

	\caption{User study for visual comparison. (a) $\{L,R\}$ vs. $\{I_M,I_M\}$; (b) $\{L,R\}$ vs. $\{L,R_{Yang}\}$; (c) $\{L,R\}$ vs. $\{L_{Yang},R\}$; (d) $\{L,R\}$ vs. $\{L_{Yang}^*, R_{Yang}^*\}$.}
	\label{table:user_study}
\end{table}

\subsection{Timing Statistics}

Our system is implemented on a PC with Xeon E5-1620 v2 CPU at 3.70GHz and 32GB RAM. No GPU is used. To compare the efficiency between our method and the existing method proposed by Yang et al.~\cite{yang2012binocular}, we compare the time of a single iteration (tone mapping and computing the energy for one output image pair) for fairness.
For an input image with resolution $800\times600$, our method takes 2.36 seconds for each iteration, while Yang et al.'s method takes 22.24 seconds.
Our method has a much better time performance than Yang et al.'s since the calculation of VDP is quite time-consuming.

\subsection{Limitations}
Most of the visual content loss happens in extremely dark or bright regions as a trade-off to preserve overall contrast. In fact, our method has the best performance in preserving local details in extremely bright regions. However, we found that the local details in extremely dark regions are still relatively unobservable even though we have preserved these local details in one view. This might be because that human perception tends to miss the local details in extremely dark regions. To make details in dark regions more apparent, we may need a more complex model to measure local contour contrast. Besides, for images that do not contain any extremely dark or bright regions, our method may generate a binocular image pair where the left and right images are similar with each other. 


\section{Conclusions}

In this paper, we propose a novel perception-based binocular tone mapping method, that can generate an optimal binocular image pair from an HDR image that presents the most visual content by designing a binocular perception strategy. In practice, we formulate the problem as an optimization where the energy function consists of three terms: contrast preservation term, detail preservation term, and fusibility term. Our method outperforms the existing method in terms of both visual and time performance.

While our current method is applied on generating a binocular image pair from a single HDR image, it is natural to extend our work to binocular tone mapping from a stereoscopic HDR image pair. We may also extend our work on generating binocular LDR video sequences from HDR video sequences, with additional exploration on temporal consistency.
\section*{Acknowledgements}
We thank all reviewers and the editor for their time and the constructive comments.
This project is supported by the Research Grants Council of the Hong Kong Special Administrative Region, under RGC General Research Fund (Project No. CUHK 14201017), 
Shenzhen Science and Technology Program (No. JCYJ20160429190300857), 
and Shenzhen Key Laboratory (No. ZDSYS201605101739178).


\bibliographystyle{eg-alpha-doi}
\bibliography{binotone}

\newcommand{\etalchar}[1]{$^{#1}$}
\begin{thebibliography}{\uppercase{HCZH{\etalchar{*}}15}}

\bibitem[AOS{\etalchar{*}}17]{alakuijala2017guetzli}
\textsc{Alakuijala J., Obryk R., Stoliarchuk O., Szabadka Z., Vandevenne L.,
  Wassenberg J.}:
\newblock Guetzli: Perceptually guided jpeg encoder.
\newblock \emph{arXiv preprint arXiv:1703.04421} (2017).

\bibitem[APH{\etalchar{*}}14]{aubry2014fast}
\textsc{Aubry M., Paris S., Hasinoff S.~W., Kautz J., Durand F.}:
\newblock Fast local laplacian filters: Theory and applications.
\newblock \emph{ACM Transactions on Graphics (TOG) 33}, 5 (2014), 167.

\bibitem[Can86]{canny1986computational}
\textsc{Canny J.}:
\newblock A computational approach to edge detection.
\newblock \emph{IEEE Transactions on Pattern Analysis \& Machine Intelligence},
  6 (1986), 679--698.

\bibitem[Cha13]{chandler2013seven}
\textsc{Chandler D.~M.}:
\newblock Seven challenges in image quality assessment: past, present, and
  future research.
\newblock \emph{ISRN Signal Processing 2013} (2013).

\bibitem[CR78]{curtis1978binocular}
\textsc{Curtis D.~W., Rule S.~J.}:
\newblock Binocular processing of brightness information: A vector-sum model.
\newblock \emph{Journal of Experimental Psychology: Human Perception and
  Performance 4}, 1 (1978), 132.

\bibitem[Dal92]{daly1992visible}
\textsc{Daly S.~J.}:
\newblock Visible differences predictor: an algorithm for the assessment of
  image fidelity.
\newblock In \emph{Human Vision, Visual Processing, and Digital Display III}
  (1992), vol.~1666, International Society for Optics and Photonics, pp.~2--16.

\bibitem[DD02]{durand2002fast}
\textsc{Durand F., Dorsey J.}:
\newblock Fast bilateral filtering for the display of high-dynamic-range
  images.
\newblock In \emph{ACM Transactions on Graphics (TOG)} (2002), vol.~21, ACM,
  pp.~257--266.

\bibitem[DWL74]{de1974binocular}
\textsc{De~Weert C.~M., Levelt W. J.~M.}:
\newblock Binocular brightness combinations: Additive and nonadditive aspects.
\newblock \emph{Perception \& Psychophysics 15}, 3 (1974), 551--562.

\bibitem[Eng69]{engel1969autocorrelation}
\textsc{Engel G.}:
\newblock The autocorrelation function and binocular brightness mixing.
\newblock \emph{Vision Research 9}, 9 (1969), 1111--1130.

\bibitem[FFLS08]{farbman2008edge}
\textsc{Farbman Z., Fattal R., Lischinski D., Szeliski R.}:
\newblock Edge-preserving decompositions for multi-scale tone and detail
  manipulation.
\newblock In \emph{ACM Transactions on Graphics (TOG)} (2008), vol.~27, ACM,
  p.~67.

\bibitem[GK99]{grossberg1999neural}
\textsc{Grossberg S., Kelly F.}:
\newblock Neural dynamics of binocular brightness perception.
\newblock \emph{Vision Research 39}, 22 (1999), 3796--3816.

\bibitem[HCZH{\etalchar{*}}15]{hau2015colorbless}
\textsc{Hau~Chua S., Zhang H., Hammad M., Zhao S., Goyal S., Singh K.}:
\newblock Colorbless: augmenting visual information for colorblind people with
  binocular luster effect.
\newblock \emph{ACM Transactions on Computer-Human Interaction (TOCHI) 21}, 6
  (2015), 32.

\bibitem[HSG{\etalchar{*}}16]{hasinoff2016burst}
\textsc{Hasinoff S.~W., Sharlet D., Geiss R., Adams A., Barron J.~T., Kainz F.,
  Chen J., Levoy M.}:
\newblock Burst photography for high dynamic range and low-light imaging on
  mobile cameras.
\newblock \emph{ACM Transactions on Graphics (TOG) 35}, 6 (2016), 192.

\bibitem[Leg84]{legge1984binocular}
\textsc{Legge G.~E.}:
\newblock Binocular contrast summation-{II}. {Q}uadratic summation.
\newblock \emph{Vision Research 24}, 4 (1984), 385--394.

\bibitem[Lev65]{levelt1965binocular}
\textsc{Levelt W.~J.}:
\newblock Binocular brightness averaging and contour information.
\newblock \emph{British Journal of Psychology 56}, 1 (1965), 1--13.

\bibitem[LR81]{legge1981binocular}
\textsc{Legge G.~E., Rubin G.~S.}:
\newblock Binocular interactions in suprathreshold contrast perception.
\newblock \emph{Perception \& Psychophysics 30}, 1 (1981), 49--61.

\bibitem[LTS92]{liu1992failure}
\textsc{Liu L., Tyler C.~W., Schor C.~M.}:
\newblock Failure of rivalry at low contrast: evidence of a suprathreshold
  binocular summation process.
\newblock \emph{Vision Research 32}, 8 (1992), 1471--1479.

\bibitem[Lub95]{lubin1995visual}
\textsc{Lubin J.}:
\newblock A visual discrimination model for imaging system design and
  evaluation.
\newblock In \emph{Vision Models for Target Detection and Recognition: In
  Memory of Arthur Menendez}. World Scientific, 1995, pp.~245--283.

\bibitem[MG05]{maehara2005binocular}
\textsc{Maehara G., Goryo K.}:
\newblock Binocular, monocular and dichoptic pattern masking.
\newblock \emph{Optical Review 12}, 2 (2005), 76--82.

\bibitem[MGB05]{meese2005interocular}
\textsc{Meese T.~S., Georgeson M.~A., Baker D.~H.}:
\newblock Interocular masking and summation indicate two stages of divisive
  contrast gain control.
\newblock In \emph{Twenty-eighth European Conference on Visual Perception}
  (2005).

\bibitem[MGB06]{meese2006binocular}
\textsc{Meese T.~S., Georgeson M.~A., Baker D.~H.}:
\newblock Binocular contrast vision at and above threshold.
\newblock \emph{Journal of Vision 6}, 11 (2006), 7--7.

\bibitem[MH04]{meese2004low}
\textsc{Meese T.~S., Hess R.~F.}:
\newblock Low spatial frequencies are suppressively masked across spatial
  scale, orientation, field position, and eye of origin.
\newblock \emph{Journal of Vision 4}, 10 (2004), 2--2.

\bibitem[MKRH11]{mantiuk2011hdr}
\textsc{Mantiuk R., Kim K.~J., Rempel A.~G., Heidrich W.}:
\newblock Hdr-vdp-2: a calibrated visual metric for visibility and quality
  predictions in all luminance conditions.
\newblock In \emph{ACM Transactions on Graphics (TOG)} (2011), vol.~30, ACM,
  p.~40.

\bibitem[PHK11]{paris2011local}
\textsc{Paris S., Hasinoff S.~W., Kautz J.}:
\newblock Local laplacian filters: Edge-aware image processing with a laplacian
  pyramid.
\newblock \emph{ACM Transactions on Graphics (TOG) 30}, 4 (2011), 68--1.

\bibitem[RGP{\etalchar{*}}04]{rizzi2004human}
\textsc{Rizzi A., Gatta C., Piacentini B., Fierro M., Marini D.}:
\newblock Human-visual-system-inspired tone mapping algorithm for hdr images.
\newblock In \emph{Human Vision and Electronic Imaging IX} (2004), vol.~5292,
  International Society for Optics and Photonics, pp.~57--69.

\bibitem[RWPD05]{reinhard2005HDR}
\textsc{Reinhard E., Ward G., Pattanaik S., Debevec P.}:
\newblock \emph{High Dynamic Range Imaging: Acquisition, Display, and
  Image-Based Lighting (The Morgan Kaufmann Series in Computer Graphics)}.
\newblock 2005.

\bibitem[SG00]{steinman2000foundations}
\textsc{Steinman B.~A., Garzia R.~P.}:
\newblock \emph{Foundations of binocular vision: a clinical perspective}.
\newblock McGraw Hill Professional, 2000.

\bibitem[SMHW16]{shen2016seamless}
\textsc{Shen W., Mao X., Hu X., Wong T.-T.}:
\newblock Seamless visual sharing with color vision deficiencies.
\newblock \emph{ACM Transactions on Graphics (TOG) 35}, 4 (2016), 70.

\bibitem[Wil17]{wilson2017binocular}
\textsc{Wilson H.~R.}:
\newblock Binocular contrast, stereopsis, and rivalry: Toward a dynamical
  synthesis.
\newblock \emph{Vision Research 140} (2017), 89--95.

\bibitem[YZWH12]{yang2012binocular}
\textsc{Yang X., Zhang L., Wong T.-T., Heng P.-A.}:
\newblock Binocular tone mapping.
\newblock \emph{ACM Transactions on Graphics (TOG) 31}, 4 (2012), 93.

\bibitem[ZW97]{zhang1997spatial}
\textsc{Zhang X., Wandell B.~A.}:
\newblock A spatial extension of cielab for digital color-image reproduction.
\newblock \emph{Journal of the Society for Information Display 5}, 1 (1997),
  61--63.

\end{thebibliography}

%
%

\end{document}